\documentclass[journal]{IEEEtran}

\usepackage[pdftex]{graphicx}
\usepackage{subfigure}
\usepackage{amsmath,amssymb}
\usepackage{multirow}
\usepackage{url}
\usepackage{cleveref}
\usepackage{color}
\usepackage{threeparttable}
\begin{document}

\title{Motion-driven Visual Tempo Learning for Video-based Action Recognition}

\author{Yuanzhong~Liu,
Junsong~Yuan,~\IEEEmembership{Fellow,~IEEE,}
and~Zhigang~Tu,~\IEEEmembership{Member,~IEEE}
\
\thanks{Yuanzhong Liu and Zhigang Tu are with the State Key Laboratory of Information Engineering in Surveying, Mapping and Remote Sensing, Wuhan University, 430079 Wuhan, China. (E-Mail: tuzhigang@whu.edu.cn, yzliu.me@whu.edu.cn). Yuanzhong Liu and Zhigang Tu contributed equally.}
\thanks{Junsong Yuan is with the Computer Science and Engineering department, State University of New York at Buffalo, USA. (Email: jsyuan@buffalo.edu)}
}

\markboth{IEEE Transactions on Image Processing}%
{Shell \MakeLowercase{\textit{et al.}}: Bare Demo of IEEEtran.cls for IEEE Journals}

\maketitle

\begin{abstract}
Action visual tempo characterizes the dynamics and the temporal scale of an action, which is helpful to distinguish human actions that share high similarities in visual dynamics and appearance. Previous methods capture the visual tempo either by sampling raw videos with multiple rates, which require a costly multi-layer network to handle each rate, or by hierarchically sampling backbone features, which rely heavily on high-level features that miss fine-grained temporal dynamics. In this work, we propose a Temporal Correlation Module (TCM), which can be easily embedded into the current action recognition backbones in a plug-in-and-play manner, to extract action visual tempo from low-level backbone features at single-layer remarkably. Specifically, our TCM contains two main components: a Multi-scale Temporal Dynamics Module (MTDM) and a Temporal Attention Module (TAM). MTDM applies a correlation operation to learn pixel-wise fine-grained temporal dynamics for both fast-tempo and slow-tempo. TAM adaptively emphasizes expressive features and suppresses inessential ones via analyzing the global information across various tempos. Extensive experiments conducted on several action recognition benchmarks, e.g. Something-Something V1 $\&$ V2, Kinetics-400, UCF-101, and HMDB-51, have demonstrated that the proposed TCM is effective to promote the performance of the existing video-based action recognition models for a large margin. The source code is publicly released at \url{https://github.com/yzfly/TCM}.
\end{abstract}


\IEEEpeerreviewmaketitle

\begin{figure*}[tbp]
  \includegraphics[width=\textwidth]{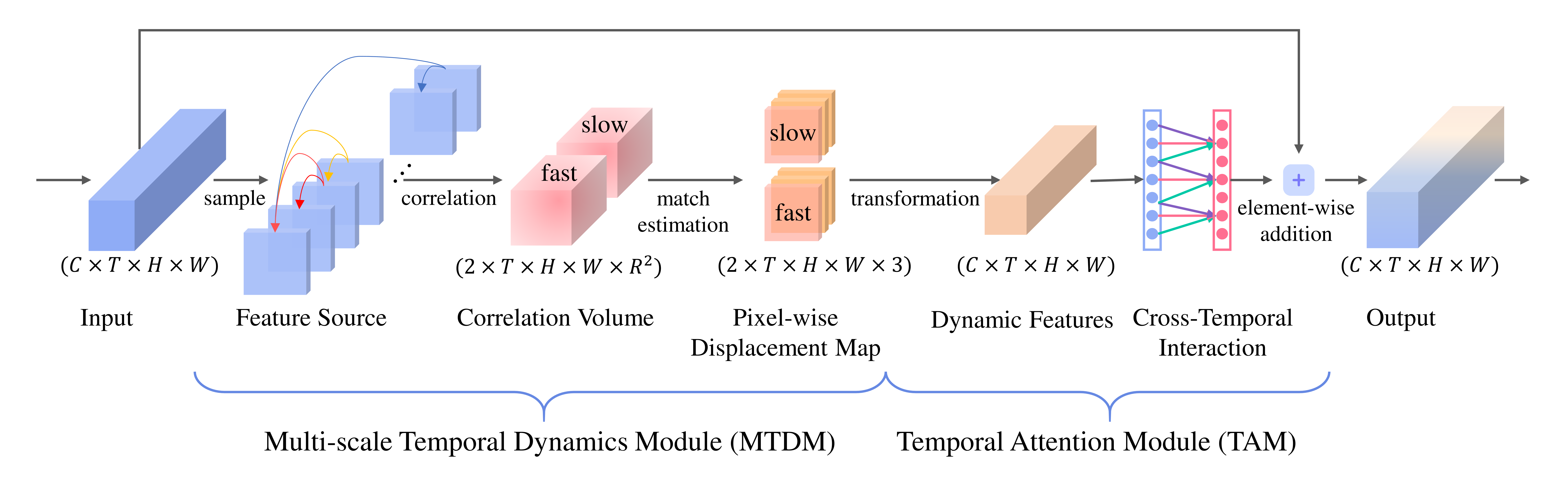}
  \caption{The architecture of our TCM. TCM includes two main components: MTDM and TAM. In MTDM, given a single-layer backbone features as \textit{Input}, we utilize a multi-scale sampling strategy to \textit{sample} the longest scale and shortest scale feature pairs for each frame to capture the \textit{Feature Source}. A \textit{correlation} is applied to both the longest scale feature pairs and the shortest scale feature pairs to form two \textit{Correlation Tensors} respectively for slow-tempo and fast-tempo. Then, we perform a \textit{motion estimation} to extract the \textit{Pixel-wise Displacement Maps} as MTDM's output, which will be fed into TAM. TAM applies \textit{transformation} on the \textit{Pixel-wise Displacement Map} to exploit \textit{Dynamic Features}. After that, a \textit{Cross-temporal Interaction} is executed to learn temporal attention weights for useful action visual tempo features excitation. Finally, the obtained action visual tempo features are combined with the \textit{Input} features as the \textit{Output}. $C, T, H, W$ respectively represents the features' channel, temporal dimension, height, and width.}
  \label{fig:arch}
\end{figure*}

\section{Introduction}
\IEEEPARstart{D}{ue} to the success of applying deep learning methods on video understanding tasks, the accuracy of video action recognition has improved significantly over the past years~\cite{simonyan2014two,t2016du,girdhar2019video,feichtenhofer2020x3d,arnab2021vivit}. However, modeling action visual tempo in videos is often overlooked. Action visual tempo describes how fast an action goes, which tends to determine the time duration at the temporal scale for recognition~\cite{yang2020temporal}. Different person performs the action at his/her own action visual tempo due to various factors e.g. age, gender, strength, and mood, etc. The complexity of action visual tempo leads to a large difference in the temporal dynamics and temporal scale. Failing to capture the action visual tempo in videos may hinder to improve the accuracy of action recognition, especially in some cases where the human actions have high similarity in dynamics and appearance (e.g., walking, jogging, and running), as distinguishing them heavily depend on extracting their action visual tempo information.

Recently, a few attempts~\cite{yang2020temporal,zhang2018dynamic,feichtenhofer2019slowfast,yang2020video} have been proposed to address this issue. SlowFast~\cite{feichtenhofer2019slowfast} samples video frames at two different rates as input to form a two-pathway SlowFast model for video recognition, where the \textit{slow} pathway operates at a low frame rate while the \textit{fast} pathway operates at a high frame rate. The backbone subnetworks accordingly aggregate the fast-tempo and slow-tempo information jointly and then handle action instances at two temporal scales. Noticeable improvements have been obtained, but this method remains computationally expensive to process action visual tempo since it uses different frame sampling rates.
Inspired by the feature-level pyramid networks~\cite{t2014Khan,t2018Cholakkal,chen2021person} which can deal with large variance in spatial scales, TPN~\cite{yang2020temporal} constructs a temporal pyramid by collecting backbone features from multi-layers and aggregating them to capture the action visual tempo information at the feature-level. TPN shows consistent improvement on several action recognition datasets, but it extremely relies on the temporal modeling ability of the backbone network itself, limiting to gain the benefits from low-level features for action recognition. 

Although the high-level features contain more semantic information, useful fine-grained temporal dynamics and temporal scale information are not fully utilized in TPN. Low-level features contain abundant fine-grained temporal dynamics and temporal scale information, thus it is not wise to sacrifice the modeling of low-level action visual tempo for improving the performance. Neglecting the process of low-level features will also result in the loss of semantic information. On the other side, high-level features are considered to contain more action semantic information since they have a large theoretical receptive field formed by deep stacks of convolutional operations~\cite{lecun1989backpropagation}. However, in fact, studies~\cite{luo2016understanding} have found that the effective receptive field in CNN only takes a small portion of the theoretical receptive field. Large effective receptive field is vital for capturing long-distance dependencies~\cite{lecun1989backpropagation,NonLocal2018}. Consequently, to enlarge the effective receptive field, it is necessary to perform the multi-scale process for low-level features directly.

Motivated by the optical flow estimation methods~\cite{tuDeepNLF,sun2018pwc,t2021liteflownet} and the motion representation-based action recognition methods~\cite{piergiovanni2019representation,kwon2020motionsqueeze,chen2020temporalc}, which estimate fine-grained motion information from low-level features to facilitate video analysis, we propose a Temporal Correlation Module (TCM) to capture the action visual tempo from the low-level features at multi-scale temporal dimension to promote the performance of current video-based action recognition models~\cite{zheng2020dynamic,ji2019context,liu2021tam}. 
As shown in Fig.~\ref{fig:arch}, TCM is composed of two parts: a Multi-scale Temporal Dynamics Module (MTDM) which is used for extracting both the slow-tempo and fast-tempo temporal dynamics, and a Temporal Attention Module (TAM) which is used for aggregating the temporal dynamics. Specifically, in MTDM, the low-level backbone features (e.g. the output features of layer \textit{res2} and layer \textit{res3} in the backbone shown in Table~\ref{tab:backbone-r50}) will serve as the feature source to establish the shortest and longest temporal feature pairs for each video frame. Then, we apply the correlation operation~\cite{dosovitskiy2015flownet} to each feature pairs to construct a correlation tensor, which is processed by an efficient motion estimation method~\cite{kwon2020motionsqueeze}, to extract pixel-wise fine-grained temporal dynamics for both fast-tempo and slow-tempo at each frame. The output of MTDM will be fed into the downstream TAM for enhancement. TAM can adaptively highlight discriminate features and reduce insignificant ones by taking the interaction of temporal dynamics at different scales into account.

Correspondingly, by equipping with the explored TCM, a powerful neural network -- TCM-Net is constructed. We integrate our TCM into the low-level layer of various action recognition backbone networks and evaluate it extensively on the popular action recognition benchmark datasets: Kinetics-400~\cite{kay2017kinetics}, HMDB-51~\cite{Kuehne11}, UCF-101~\cite{soomro2012ucf101}, and Something-Something V1 $\&$ V2~\cite{goyal2017something}. The experimental results show that the prior action recognition methods can achieve impressive gains when combine with our TCM. As pointed in~\cite{jiang2019stm, li2020tea,lin2019tsm}, most of human actions cannot be recognized in the temporal dominated videos without considering the temporal relationship, like the human actions in the Something-Something V1 $\&$ V2 video datasets. Specifically, when incorporating TCM into the basic backbone ResNet50~\cite{he2016deep} (TCM is placed right behind layer $res3$, see Table \ref{tab:backbone-r50} for layer reference), the modified TCM-R50 model (with only 4\% more FLOPs than ResNet50) produces competitive result, which is on par with the prior best performance on the Something-Something V1 $\&$ V2 datasets and the Kinetics400 dataset. Besides, a comprehensive ablation studies also demonstrate the effectiveness and efficiency of the two components of TCM \textit{i.e.} MTDM and TAM.

Our main contributions are summarized as follows:
\begin{itemize}
\item We design a MTDM to fully extract the pixel-wise fine-grained temporal dynamics of both fast-tempo and slow-tempo from the low-level single-layer deep features, which addresses the limitations of the previous methods that heavily rely on high-level features and are unable to exploit benefits from low-level features.
\item We exploit a TAM, which can adaptively select and enhance the most effective action visual tempo from multi-scale temporal dynamics, to aggregate temporal dynamics.
\item A TCM is constructed by combining MTDM with TAM, which can incorporate with various action recognition backbone networks easily in a plug-in-and-play way. Extensive experiments conducted on the main 2D and 3D action recognition backbones and action recognition benchmarks show that our TCM significantly improves the accuracy of current video-based action recognition models.
\end{itemize}

\section{Related Work}
\subsection{Action Recognition in Videos}
The current convolutional deep learning methods~\cite{yang2020sta,ji2019context,tu2018semantic} dedicated to human action recognition can be roughly divided into two categories, \textit{i.e.} 3D convolutional networks (3D CNNs) and 2D convolutional networks (2D CNNs). 3D CNNs~\cite{tran2015learning,hara2018can,t2017Sabokrou,feichtenhofer2020x3d} utilize 3D convolutional kernels to jointly model temporal and spatial semantics. Local temporal convolution operations are stacked to capture the long-range temporal dynamics. The Non-local network~\cite{NonLocal2018} introduces a non-local operation to better exploit the long-range temporal dynamics from input sequences. Except the non-local operation, there are many other modifications~\cite{carreira2017quo,tran2018closer,piergiovanni2019representation} have been explored to 3D CNNs to boost its performance, but the variation of action visual tempo is often neglected. 2D CNNs~\cite{wang2016temporal,zhou2018temporal, lin2019tsm,t2021res2net} apply 2D kernels over per-frame inputs to exploit spatial semantics and followed by a module to aggregate temporal dynamics. The most famous 2D CNN model is the two-steam network~\cite{simonyan2014two,wang2016temporal,liu2019ntu}, in which one stream extracts the RGB appearance features, and the other stream learns the optical flow motion information. Finally, it uses the average pooling for spatio-temporal aggregation. A number of efforts~\cite{jiang2019stm,tu2019action,li2020tea,kwon2020motionsqueeze} has been carried out to enhance the temporal information extraction efficiency for 2D CNNs. STM~\cite{jiang2019stm} proposes a Channel-wise Spatiotemporal Module and a Channel-wise Motion Module to encode the complementary spatiotemporal and motion features in a unified 2D CNN framework.
ActionS-ST-VLAD~\cite{tu2019action} propose a novel action-stage(ActionS) emphasized spatiotemporal vector of locally aggregated descriptors (ActionS-ST-VLAD) method to adaptively aggregate video-level informative deep features.
TEA~\cite{li2020tea} calculates the feature-level temporal differences from spatiotemporal features and utilizes the differences to excite the motion-sensitive channels of the features. MotionSqueeze~\cite{kwon2020motionsqueeze} presents a trainable neural module to establish correspondence across frames and convert them into motion features. These methods provide fine-grained modeling ability to learn adjacent frame temporal dynamics, but they ignore the importance of action visual tempo.

\subsection{Action visual tempo Modeling in Video}
Many methods~\cite{yang2020temporal,zhang2018dynamic,feichtenhofer2019slowfast} are dedicated to action visual tempo~\cite{wang2008visual} dynamics modeling by taking advantages of the input-level frame pyramid. DTPN~\cite{zhang2018dynamic} samples video frames with varied frame sampling rates and constructs a pyramidal feature representation for arbitrary-length input videos, where the slow-tempo and fast-tempo temporal dynamics can be captured. Such sample strategy tends to require multiple frames which causes a heavy computational cost, especially when the frame sampling rate increases. SlowFast~\cite{feichtenhofer2019slowfast} uses a two-level frame pyramid to hard-code the variance of the action visual tempo.
Branches are carefully devised to separately process each level, and the mid-level features of these branches are fused interactively. SlowFast can robustly deal with the variance of the action visual tempo, but the multi-branch network is costly. TPN~\cite{yang2020temporal} leverages different depth features hierarchically formed inside the backbone network to model action visual tempo. TPN can be applied to various models and brings consistent improvement, but it is limited by the backbone networks' temporal modeling ability. It cannot take advantage of the low-level features, and the useful long-range fine-grained temporal dynamics from distant frames in high-level features may be weakened as it is obtained by stacking multiple local temporal convolutions. To overcome these disadvantages, we utilize the correlation operation to establish pixel-wise matching values for different temporal-scale features and exploit the action visual tempo in videos. We also explore a temporal attention module for interactive action visual tempo feature fusion, which not only enhances the saliency temporal scales, but also enriches the temporal dynamics.

\section{Proposed Method}
In this section, we will introduce the details of our proposed TCM (see Fig. \ref{fig:arch}). TCM contains two important parts: a Multi-scale Temporal Dynamics Module (MTDM) and a Temporal Attention Module (TAM). Initially, the visual contents of the input video are encoded into a feature sequence by a spatio-temporal action recognition backbone network. Then, the designed MTDM utilizes a correlation operation to extract the temporal dynamics of both fast-tempo and slow-tempo from this feature sequence. To determine the most effective visual tempo information, TAM will adaptively emphasize expressive temporal features and suppress insignificant ones by analyzing across-temporal interactions. Lastly, the obtained action visual tempo features are combined with the appearance features for final prediction.

\begin{figure*}[tbp]
\subfigure[fast-tempo]{
\label{Fig.sub.f}
\includegraphics[width=0.3\textwidth]{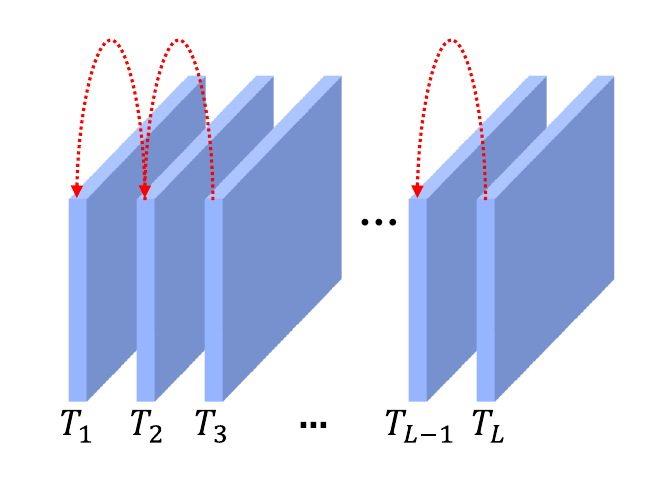}}
\subfigure[slow-tempo]{
\label{Fig.sub.s}
\includegraphics[width=0.3\textwidth]{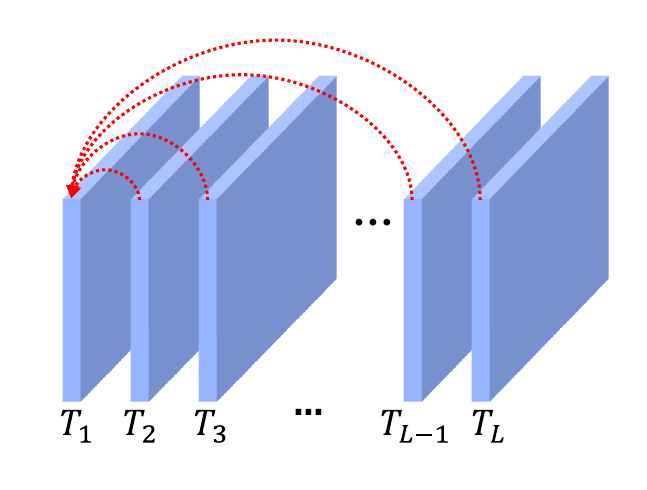}}
\subfigure[slow-fast-tempo]{
\label{Fig.sub.fs}
\includegraphics[width=0.35\textwidth]{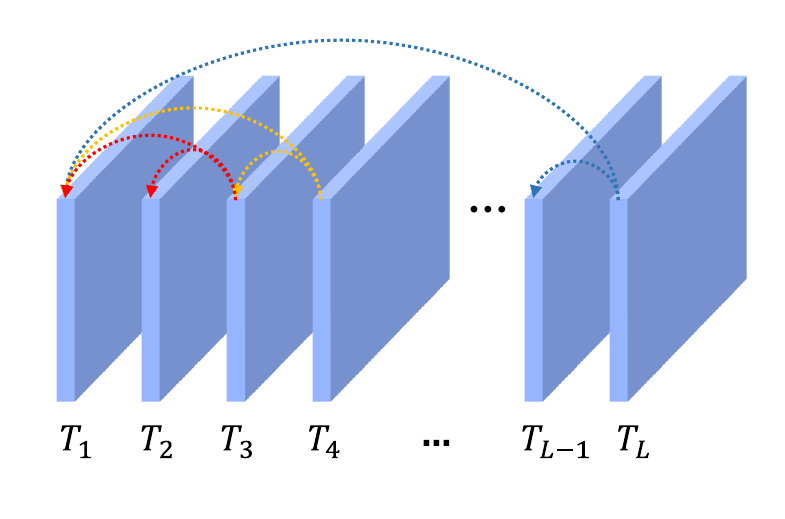}}
\caption{Slow-fast-tempo Feature Sources. The longest scale and the shortest scale are considered for each moment to form a multi-scale sampling strategy to model the slow-fast-tempo information. The longest scale has a large temporal receptive field for slow-tempo, and the shortest scale has a small temporal receptive field for fast-tempo.}
\label{fig:corr}
\end{figure*}

\subsection{Multi-scale Temporal Dynamics Module (MTDM)}
The MTDM is a learnable temporal dynamics extractor, which extracts effective temporal dynamic features of both the fast-tempo and the slow-tempo in three steps: feature source utilization, visual similarity computation, and motion estimation.

\subsubsection{Feature Source Utilization}
The prior feature-based methods~\cite{yang2020temporal} utilize the high-level backbone features to construct a multi-layer feature pyramid that has increasing temporal receptive fields from bottom to top. The single-layer pyramid feature source has also been explored~\cite{yang2020temporal}, but the effectiveness is limited due to the constraint of the backbone's temporal modeling ability. In contrast, we design a novel approach to extract the action visual tempo features of both the slow-tempo and the fast-tempo, where they are obtained by sampling deep features at different rates. This approach can effectively use the single-layer deep features and be free from the constraints of the backbone network.

Specifically, to learn features of the fast-tempo at each frame, we present an adjacent video frames feature extraction strategy (see Fig. \ref{Fig.sub.f}), which can effectively exploit the shortest scale temporal information. On the other side, to capture features of the slow-tempo for each moment, we use the longest scale temporal information and accordingly construct the feature extraction strategy as shown in Fig.~\ref{Fig.sub.s}. Combining the fast-tempo (Fig.~\ref{Fig.sub.f}) with the slow-tempo (Fig.~\ref{Fig.sub.s}) feature sampling strategy, our final multi-scale slow-fast-tempo sampling strategy (Fig. \ref{Fig.sub.fs}) is formed. Accordingly, for each frame, we have its shortest and longest temporal range information, \textit{i.e.} the fast-tempo and the slow-tempo. Inspired by the way to estimate optical flow~\cite{tu2019survey,t2020Young}, we calculate the pixel-wise visual similarity characteristics of each frame temporally to model the action visual tempo structure.

\begin{figure*}[tbp]
  \includegraphics[width=\textwidth]{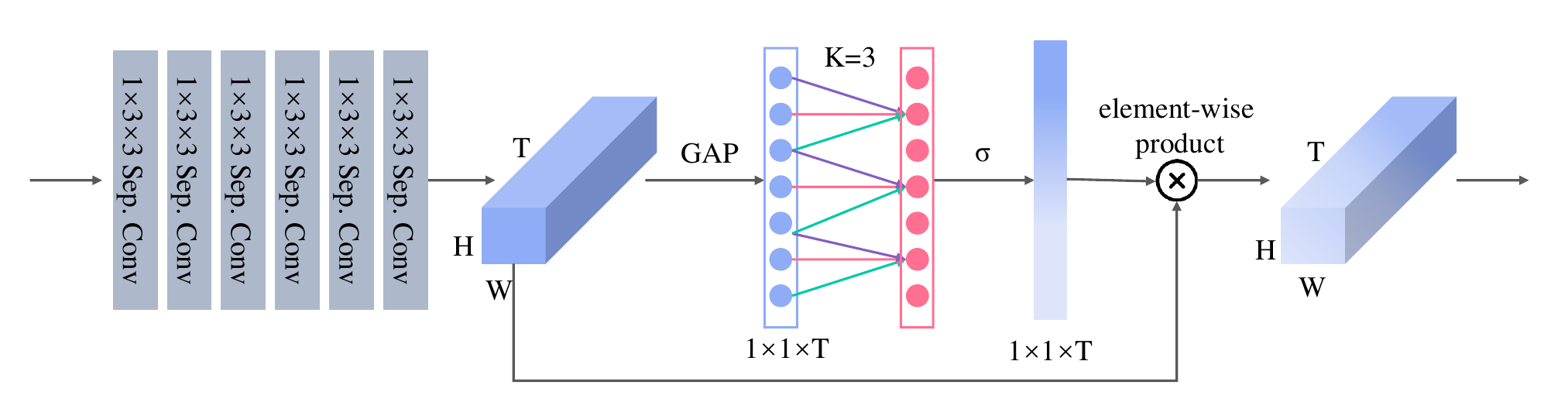}
  \caption{Our explored Temporal Attention Module (TAM). The displacement map (with confidence map) from MTDM are transformed by six convolution layers to interpret slow-fast tempo semantics. Then the global average pooling (GAP) is used to capture aggregated features. With consideration of the cross-temporal interaction, temporal weights are generated by performing a fast 1D temporal convolution with size $K$, where $K$ is adaptively determined by Eq. \ref{eq:k}, here we show the case of $K=3$.}
  \label{fig:tam}
\end{figure*}

\subsubsection{Visual Similarity Computation}
Let us denote the pair of the input feature maps at a certain interval $r$ by $F^{t}\in \mathbb{R}^{C \times H \times W}$ and $F^{t+r}\in \mathbb{R}^{C \times H \times W}$, where $C,H$ and $W$ are respectively the channel dimension, height and width. The visual similarity score at a position $\mathbf{x}$ with respect to the displacement $\mathbf{p}$ can be defined as:
\begin{equation}
    \label{eq:s}
  S(\mathbf{x},\mathbf{p},t) = F^{t}_{\mathbf{x}} \cdot F^{t+r}_{\mathbf{x}+\mathbf{p}},
\end{equation}
where $ \cdot $ represents dot product. To improve the efficiency, we compute the visual similarity score at the position $\mathbf{x}$ only in its neighborhood with the radius $R$ \textit{i.e.} $\mathbf{p} \in [-R, R]^2$.

Furthermore, to form the correlation tensor, we calculate the visual similarity score of the input feature map pairs $F^t$ and $F^{t+r}$ at each position, which can be computed as following:
\begin{equation}\label{c2}
\begin{aligned}
  & \mathbf{C}(F^{t}, F^{t+r}) \in \mathbb{R}^{H \times W \times R \times R}, \\
  & C_{i j k l;|k-i| \le R,|l-j| \le R }=\sum_{h} { {F^{t}}_{h i j } \cdot {F^{t+r}}_{h k l} }
\end{aligned}
\end{equation}
The radius $R$ is set manually and affects the final performance. In practice, given a feature map with the spatial resolution $H \times W$ (H is generally equal to W), we can set $R=INT(H/2)$, where $INT$ is the integer ceiling function. In theory, the result produced by the correlation is four-dimension ($H \times W \times R \times R$), we resize it to $H \times W \times R^2$ to facilitate subsequent processing.

\subsubsection{Motion Estimation}
To establish the correspondence across frame pairs, we use a light-weight method as MotionSqueeze~\cite{kwon2020motionsqueeze} to estimate optical flow in terms of the correlation tensor $C(F^{t}, F^{t+r})$. Following the setting of MotionSqueeze~\cite{kwon2020motionsqueeze}, we compute optical flow and it's corresponding confidence map for motion information extraction. We find that the confidence map is very useful for identifying motion outliers and learning informative motion features.
The two-channels' optical flow and the one-channel' confidence map are concatenated to form a pixel-wise displacement map. To maintain the temporal dimension consistency, we simply duplicate the last temporal dimension of the displacement map. The displacement map contains motion information of specific scale. After performing motion estimation of the longest and shortest scales, we concatenate them together for downstream feature transformation of both the slow-tempo and the fast-tempo.

\subsection{Temporal Attention Module (TAM)}
Our designed TAM aims to exploit the temporal dynamics of the upstream input, and adaptively highlight the distinctive features while suppressing the trivial ones by taking across-temporal dynamics interaction into account. It is known that the \textit{depth-wise separable convolution}~\cite{t2020Zhang} can significantly reduce the computational complexity of CNNs, and we use it to improve the efficiency here. As shown in Fig. \ref{fig:tam}, the displacement map (with the confidence map) from upstream layers will be fed into six convolution layers for transformation. Three $1 \times 3 \times 3$ layers are utilized to approximate a $1\times 7 \times 7$ layer, then followed by another three $1 \times 3 \times 3$ layers. Specifically, except for the first two layers, all the other layers are followed by a $1 \times 1$ convolution layer. The semantics of the displacement map and the confidence map are expected to be interpreted by the feature transformation.
After transforming the displacement map, the slow-fast visual tempo features are obtained for aggregation. The TAM is designed to automatically extrude the discriminative action visual tempo features meanwhile to reduce the impact of inessential features during training. Recently, there are some effective attempts to enhance the temporal information by utilizing the attention mechanism. TEA~\cite{li2020tea} employs a global average pooling layer to summarize the spatial information to get attentive weights to stimulate the motion-sensitive channels. Motion pattern has been excited and enhanced, but processing the temporal channels in isolation can lead to the loss of cross-temporal interaction.

Given the temporal aggregated feature $F^{T} \in \mathbf{R}^{T}$, where $T$ denotes the feature temporal dimension, the temporal attention can be learned without dimensionality reduction according to Equation \ref{eq:attention}:
\begin{equation}\label{eq:attention}
  w = \sigma(WF^{T}),
\end{equation}
where $W$ is a general parameter matrix with $T \times T$ elements. Specifically, the parameter matrix in TEA~\cite{li2020tea} is computed according to Equation \ref{eq:weights}:
\begin{equation}\label{eq:weights}
  W_{1} =
  \begin{bmatrix}
      w^{1,1} & \cdots & 0 \\
      \vdots  & \ddots & \vdots \\
        0     & \cdots & w^{T,T},
  \end{bmatrix}
\end{equation}
where $W_{1}$ is a diagonal matrix contains $T$ parameters, but the cross-temporal interaction is completely ignored here. Recent research about attention mechanism~\cite{Wang_2020_CVPR_eca} suggests that the cross-channel interaction is useful, and the temporal interaction has a latent important function for video analysis tasks.

We explore a novel way to capture the local cross-temporal interaction. Same as the correlation operation, when calculating the visual similarity, we convert the global operation to a local operation to improve the efficiency and accuracy. This means the weight of the temporal aggregated feature $F_{i}$ is calculated by only considering the temporal interaction with its $k$ neighbors as Eq. \ref{eq:wi}:\begin{equation}\label{eq:wi}
  w_i = \sigma(\sum_{j=1}^{k}w^{j}_{i}F_{i}^{j}), F_{i}^{j}\in \Omega _{i}^{k},
\end{equation}
where $\Omega _{i}^{k}$ indicates the set of $k$ adjacent temporal features of $F_{i}$. Accordingly, a band matrix $W_{k}$ is employed to learn the temporal attention, where $W_{k}$ is computed as:
\begin{equation}\label{eq:wvar2}
\begin{aligned}
  W_{k} = \\
  \begin{bmatrix}
      w^{1,1} & \cdots  & w^{1,k} & 0         & 0       & \cdots        & \cdots & 0 \\
      0       & w^{2,2} & \cdots  & w^{2,k+1} & 0       & \cdots        & \cdots & 0 \\
      \vdots  & \vdots  & \vdots  & \vdots    & \ddots  & \vdots        & \vdots & \vdots \\
      0       & \cdots  & 0       & 0         & \cdots  & w^{T,T-k+1}   & \cdots & w^{T,T}
  \end{bmatrix}
\end{aligned}
\end{equation}

Notably, $W_{k}$ involves only $k \times T$ parameters, which are less than $T \times T$. Besides, we make all temporal channels to share the same learning parameters to boost the calculating speed:
\begin{equation}\label{eq:wishare}
  w_i = \sigma(\sum_{j=1}^{k}w^{j}F_{i}^{j}), F_{i}^{j}\in \Omega _{i}^{k}
\end{equation}
The range of the temporal interaction (\textit{i.e.} the size of $k$) needs to be determined carefully. Following~\cite{Wang_2020_CVPR_eca}, the value of $k$ can be decided by:
\begin{equation}\label{eq:k}
  k = \phi(T) = \frac{1}{\gamma}|\log_{2}(T) + b|_{odd},
\end{equation}
where $|v|_{odd}$ indicates the nearest odd number of $v$. We set both $\gamma$ and $b$ to 1 in our experiments.

Since the current frame and its adjacent frames have both the longest and shortest scales at the same time, the current temporal receptive field is further expanded. As a result, the features have learned the information of both the fast-tempo and the slow-tempo. Taking the interaction of the temporal dynamics at different temporal scales into account, our temporal attention module can better enhance the useful slow and fast visual tempo information and suppress the unnecessary ones.

\subsection{Implementation}
In MTDM, we first apply a $1\times 1$ convolution to reduce channels to boost the computational efficiency.
The C++/Cuda implemented version of the correlation operation in FlowNet~\cite{dosovitskiy2015flownet} is adopted for our correlation tensor calculation.
The motion estimation method of~\cite{kwon2020motionsqueeze} is introduced to estimate the displacement map from the correlation tensor.
In TAM, six $1\times 3 \times 3$ depth-wise separable convolutions are used to exploit fine-grained multi-scale temporal semantics. For the temporal attention, it can be performed by a fast 1D convolution with a kernel size $k$, and then we extend the channel attention method ECA~\cite{Wang_2020_CVPR_eca} to the temporal dimension.
We utilize ResNet50~\cite{he2016deep} as the backbone, whose structure is presented in Table \ref{tab:backbone-r50}.

\begin{table}[!ht]
\centering
\caption{The illustration of our used 2D backbone ResNet50. Note that both the kernel size and the output size are in $W \times H$.}
\label{tab:backbone-r50}
\begin{tabular}{c|c|c}
\hline
Stage & Layer & Output size \\ \hline
raw & $-$ & $ 224 \times 224$ \\ \hline
conv$_{1}$ & $ 7 \times 7,64,$ stride 2,2 & $112 \times 112$ \\ \hline
pool$_{1}$ & $ 3 \times 3$ max, stride 2,2 & $56 \times 56$ \\ \hline
res$_{2}$ & {$\left[\begin{array}{c}1 \times 1,64 \\
 3 \times 3,64 \\
 1 \times 1,256\end{array}\right] \times 3$} & $56 \times 56$ \\ \hline
res$_{3}$ & {$\left[\begin{array}{l} 1 \times 1,128 \\
3 \times 3,128 \\
 1 \times 1,512\end{array}\right] \times 4$} & $28 \times 28$ \\ \hline
res$_{4}$ & {$\left[\begin{array}{c}1 \times 1,256 \\
 3 \times 3,256 \\
1 \times 1,1024\end{array}\right] \times 6$} & $14 \times 14$ \\ \hline
res$_{5}$ & {$\left[\begin{array}{c} 1 \times 1,512 \\
 3 \times 3,512  \\
1 \times 1,2048\end{array}\right] \times 3$} & $7 \times 7$ \\ \hline
\multicolumn{2}{c|}{ global average pool, fc } &  $1 \times 1$ \\ \hline
\end{tabular}
\end{table}

\section{Experiments}
We evaluate the proposed method on various action recognition datasets, including Kinetics-400~\cite{kay2017kinetics}, HMDB-51~\cite{Kuehne11}, UCF-101~\cite{soomro2012ucf101}, and Something-Something V1 \& V2~\cite{goyal2017something}. The \textit{baseline method} in our experiments is \textit{TSM}~\cite{lin2019tsm} which uses ResNet50~\cite{he2016deep} without non-local modules~\cite{NonLocal2018}, thus it is fair for comparison. Furthermore, we test our method on multiple action recognition backbone networks (\textit{i.e.} TSM, TEA, and I3D), and conduct plenty of ablation studies about the components of TCM on Something-Something V1, to analyze the effectiveness of the proposed TCM and its two components \textit{i.e.} MTDM and TAM. It should be noted that we focus on the gain of action recognition from the extraction of action visual tempo patterns, and we only use RGB frames rather than optical flow to save the computation cost.

\noindent \textbf{Datasets.}
As mentioned in the previous work~\cite{jiang2019stm}, the primary public datasets for action recognition can be roughly classified into two categories: (1) the temporal-related datasets \textit{e.g.} Something-Something V1 $\&$ V2~\cite{goyal2017something}, in which the temporal motion interaction of objects should be emphasized for better action understanding. (2) The scene-related datasets \textit{e.g.} Kinetics-400~\cite{kay2017kinetics}, UCF-101~\cite{soomro2012ucf101} and HMDB-51~\cite{Kuehne11}), in which the temporal relation is less important compared to the temporal-related datasets, this is because the background information contributes more for determining the action label in most of the videos. The Something-Something V1 \& V2 datasets focus on human interactions with daily life objects, thus classifying these interactions required to pay more attention to the temporal information. Consequently, the proposed method is mainly evaluated on Something-Something V1 \& V2 as our goal is to improve the temporal modeling ability. Additionally, we also report experimental results on the scene-related datasets Kinetics-400~\cite{kay2017kinetics}, HMDB-51~\cite{Kuehne11}, and UCF-101~\cite{soomro2012ucf101}. Kinetics-400 contains 400 human action categories, and provides $~240k$ training videos and $~20k$ validation videos. In our experiments, due to some videos in Kinetics-400 are unavailable, we collected 238,798 videos for training and 19,852 videos for validation.

\noindent \textbf{Training.} In general, we adopted the training strategy same as TSM~\cite{lin2019tsm}. Our model is initialized with the ImageNet pre-trained weights of ResNet50 (see Table \ref{tab:backbone-r50}). The training settings for the Kinetics-400, UCF-101, and HMDB-51 datasets are also the same as TSM~\cite{lin2019tsm}. For the Something-Something V1 \& V2 datasets, the training parameters are: the epochs are 50, the batch size is 32, the initial learning rate is 0.01 (decays by 0.1 at epoch 30, 40 and 45), the weight decay is 5e-4, and the dropout is 0.5. At training, for each video, we sample a clip with 8 or 16 frames, resize them to the scale of $240 \times 320$, and then crop a $224 \times 224$ patch from the resized images. The scale jittering is used for data augmentation. The final prediction follows the standard protocol of TSN~\cite{wang2016temporal}.

\noindent \textbf{Evaluation.} For the Something-Something V1 \& V2 datasets, two kinds of testing schemes are used: 1) single-clip and center-crop, where only a center crop of $224 \times 224$ from a single clip is utilized for evaluation; 2) 10-clip and 3-crop, where three crops of $224 \times 224$ and 10 randomly-sampled clips are employed for testing. The first testing scheme is with high efficiency while the second one is for improving the accuracy with a denser prediction strategy. We evaluate both the single clip prediction and the average prediction of 10 randomly-sampled clips. For the Kinetics-400 dataset, we evaluate the average prediction of uniformly-sampled 10 clips from each video. For the UCF-101 and HMDB-51 datasets, 2 uniformly-sampled clips from each video are selected for evaluation.

\begin{table}[!ht]
\centering
\caption{Comparison of our method ``TSM+TCM" with TSM on different datasets. Specifically, 8 frames are input for training. At testing, 10 video-clips for Kinetics-400, 2 video-clips for HMDB-51 and UCF-101, and a single video clip for Something-Something V1 \& V2.}
\label{tab:baseline}
\begin{tabular}{ccccc}
\hline
Dataset & Model & Top-1(\%) & Top-5(\%) & $\Delta$ Top-1(\%) \\ \hline
\multirow{2}{*} { Kinetics-400 } & TSM & 74.1 & 91.2 & \multirow{2}{*} {+1.5} \\
                                  & Ours & $\mathbf{75.6}$ & $\mathbf{92.5}$ &  \\ \hline
\multirow{2}{*} { UCF-101 } & TSM & 95.9 & 99.7 & \multirow{2}{*} {+1.3} \\
                                  & Ours & $\mathbf{97.2}$ & $\mathbf{99.8}$ &  \\ \hline
\multirow{2}{*} { HMDB-51 } & TSM & 73.5 & 94.3 & \multirow{2}{*} {+4.1} \\
                                  & Ours & $\mathbf{77.6}$ & $\mathbf{96.4}$ &  \\ \hline \hline
Sth-Sth  & TSM & 47.3 & 76.2 & \multirow{2}{*} {+4.7} \\
V1 & Ours & $\mathbf{52.0}$ & $\mathbf{80.4}$ & \\ \hline
Sth-Sth & TSM & 61.7 & 87.4 & \multirow{2}{*} {+1.7} \\
V2 & Ours & $\mathbf{63.4}$ & $\mathbf{88.6}$ & \\ \hline
\end{tabular}
\end{table}

\subsection{Performance on CNN baselines}
TCM can be seamlessly injected into a CNN baseline to significantly enhance its temporal information modeling ability. To demonstrate that the enhancement is generalized and steady, we compare our TCM with some baseline networks on some famous action recognition benchmarks.

\subsubsection{Evaluation on Different Datasets} In this experiment, as analyzed before, we select the representative model TSM~\cite{lin2019tsm} as the CNN baseline, and use the same training and testing protocols for both the original TSM~\cite{lin2019tsm} and the modified model ``TSM+TCM" for fair comparison.
The results are shown in Table \ref{tab:baseline}. In the upper part, for the datasets Kinetics-400, UCF-101, and HMDB-51, their temporal information is relatively less important. In contrast, in the lower part, for the datasets Something-Something V1 \& V2, the temporal information becomes very important. When integrating our TCM module into TSM, the performance of ``TSM+TCM" has significantly improved on both the scene dominant datasets and the temporal dominant datasets. For example, compared with TSM, on the large-scale dataset Kinetics-400, the Top-1 accuracy of ``TSM+TCM" is improved by 1.5\% (75.6\% vs. 74.1\%); On the relatively smaller-scale datasets UCF-101 and HMDB-51, the Top-1 accuracy of ``TSM+TCM" is boosted respectively by 1.3\% and 4.1\%; On the temporal dominated datasets Something-Something V1 \& V2, the performance improvement is more obvious, where the Top-1 accuracy is enhanced separately by 4.7\% and 1.7\%. This proves that the proposed TCM is effective to improve the temporal modeling ability of the baseline.

\begin{table*}[!ht]
\centering
\caption{Comparison between our TCM and other backbones on the Something-Something V1 dataset.}
\label{tab:diff_backbones}
\begin{tabular}{lccclllll}
\hline
                      & Method       & TCM          & Input & FLOPs            & Params   & \multicolumn{2}{c}{Sth-Sth V1}  \\
                             &                &              &       &                  &         & Top-1(\%)  & Top-5(\%)         \\ \hline
\multirow{8}{*}{2D}  & \multirow{2}{*}{TSM-ResNet18~\cite{wang2016temporal} }      & $\times$         & $224\times224\times8$     & 14.5G    & 11.3M   & 42.8           & 72.3                \\
                             &                & $\checkmark$ & $224\times224\times8$   & 16.4G$_{(+1.9)}$    & 11.5M$_{(+0.2)}$   & 45.8$_{(+3.0)}$  & 74.8$_{(+2.5)}$ \\ \cline{2-8}
                             & \multirow{2}{*}{TSM-ResNet50~\cite{wang2016temporal}}      &  $\times$       & $224\times224\times8$     & 33.4G    & 24.3M   & 45.6           & 74.2                \\
                             &                & $\checkmark$ & $224\times224\times8$   & 35.3G$_{(+1.9)}$    & 24.5M$_{(+0.2)}$   & 52.0$_{(+6.4)}$  & 80.4$_{(+6.2)}$ \\ \cline{2-8}
                             & \multirow{2}{*}{TSM-ResNet101~\cite{wang2016temporal}}     &  $\times$   & $224\times224\times8$     & 63.1G    & 42.9M   & 46.3           & 75.8                \\
                             &                & $\checkmark$ & $224\times224\times8$   & 65.0G$_{(+1.9)}$    & 43.1M$_{(+0.2)}$   & 52.6$_{(+6.3)}$  & 81.4$_{(+5.6)}$ \\ \cline{2-8}
  & \multirow{2}{*}{TEA~\cite{zhou2018temporal}}     &$\times$   & $224\times224\times8$     & 34.7G    & 24.4M      & 48.9       &  78.1               \\
                             &                & $\checkmark$ & $224\times224\times8$     & 36.6G$_{(+1.9)}$  & 24.6M$_{(+0.2)}$   & 50.6$_{(+1.7)}$   & 79.7$_{(+1.6)}$  \\ \hline
\multirow{8}{*}{3D}
&\multirow{2}{*}{3D-ResNet-18~\cite{hara2018can}}    & $\times$     & $112\times112\times16$    & 33.2G    & 33.3M    & 16.4       & 45.3                        \\
                             &                & $\checkmark$ & $112\times112\times16$     & 33.4G$_{(+0.2)}$    & 33.5M$_{(+0.2)}$   &19.7$_{(+3.3)}$   & 48.5$_{(+3.2)}$    \\ \cline{2-8}
&\multirow{2}{*}{3D-ResNet-50~\cite{hara2018can}} &$\times$     & $112\times112\times16$    & 40.4G    & 46.2M    & 22.3       & 51.4                        \\
                             &                & $\checkmark$ & $112\times112\times16$     & 40.6G$_{(+0.2)}$    & 46.4M$_{(+0.2)}$   &24.1$_{(+1.8)}$   & 53.7$_{(+2.3)}$    \\ \cline{2-8}
&\multirow{2}{*}{3D-ResNeXt from~\cite{crasto2019mars}}  &$\times$     & $112\times112\times16$    & 9.6G    & 47.8M    & 24.5       & 53.2                        \\
                             &                & $\checkmark$ & $112\times112\times16$     & 9.8G$_{(+0.2)}$    & 48.2M$_{(+0.2)}$   &26.1$_{(+1.6)}$   & 55.7$_{(+2.5)}$    \\ \hline

\end{tabular}
\end{table*}

\subsubsection{Evaluation on Different Backbones} We apply our TCM to a variety of backbone networks, and show their accuracy on the Something-Something V1 dataset in Table \ref{tab:diff_backbones}. There are two parts: the upper part is 2D-CNN methods, and the lower part is 3D-CNN methods. Particularly, in all networks, TCM is placed right behind layer \textit{res3}. For TSM over different backbones, ``TSM+TCM" can clearly enhance the accuracy of action recognition with only a small increase in parameters. Compared to the baseline TSM-ResNet50, TCM obtains a significant gain of about 6.4\% (52.0\% vs. 45.6\%) at Top-1 accuracy at the cost of only 5.6\%  (35.3G vs. 33.4G) and 0.8\% (24.5M vs. 24.3M) growth in FLOPs and parameters. For the well-performed TEA~\cite{li2020tea} which can stimulate and aggregate the temporal information effectively, our TCM also boosts its performance for a large margin, \textit{e.g.} the Top-1 accuracy enhances from 48.9\% to 50.6\%. For 3D-ResNet~\cite{hara2018can} over different depth, TCM can steadily promote the performance, \textit{e.g.} 3D-ResNet-18 (Top-1 + 3.3\%), 3D-ResNet-50 (Top-1 + 1.8\% ). For the famous 3D network 3D-ResNeXt~\cite{crasto2019mars}, our TCM further improves its temporal modeling capability, where the Top-1 accuracy is modified by 1.6\% (24.5\% vs. 26.1\%).

\begin{table*}
\caption{Performance comparison with state-of-the-arts on the Something-Something V1 \& V2 datasets. Most of the results are copied from the corresponding paper, and the symbol ``-” denotes the result is not given.}
\label{tab:sota}
\centering
\begin{tabular}{l|c|c|c|c|c|c|c}
\hline
Method                          & Frame & FLOPs $\times$ clips      & Params & \multicolumn{2}{c}{Sth-Sth V1} & \multicolumn{2}{c}{Sth-Sth V2} \\
                                      &       &                           &       & Top-1(\%)          & Top-5(\%)          & Top-1(\%)          & Top-5(\%)  \\ \hline
ECO$_{En}Lite$~\cite{zolfaghari2018eco}       & 92    & 267 $\times$ 1            & 150M  & 46.4           & -              & -              & -              \\
I3D from~\cite{wang2018videos}            & 32    & 153G $\times$ 2  & 28.0M & 41.6           & 72.2           & -              & -              \\
NL-I3D from~\cite{wang2018videos}         & 32    & 168G $\times$ 2  & 35.3M & 44.4           & 76.0           & -        & -              \\
NL-I3D + GCN~\cite{wang2018videos}    & 32    & 303G $\times$ 2  & 62.2M & 46.1           & 76.8           & -         & -              \\
S3D-G~\cite{xie2018rethinking}   & 64    & 71G $\times$ 1   & 11.6M & 48.2           & 78.7           & -              & -              \\
DFB-Net~\cite{martinez2019action}   & 16    & N/A $\times$ 1   & -     & 50.1           & 79.5           & -              & -              \\
CorrNet-101~\cite{wang2020video}   & 32    & 224G $\times$ 30   & -     & 51.7           & -           & -              & -              \\
3D DenseNet121~\cite{zhou2020spatiotemporal}  & 16    & 31G $\times$ 1   & 21.4M     & 50.2           & 78.9           & 62.9              & 88.0              \\
CIDC(I3D)~\cite{li2020directional}  & 32    & 92G $\times$ 30   & 87M     & -           & -           & 56.3              & 83.7              \\
RubiksNet~\cite{fan2020rubiksnet}   & 8    & 33G $\times$ 1   & -     & 46.4           & 74.5           & 58.8              & 85.6              \\
\hline
TSN~\cite{wang2016temporal}   & 8     & 16G $\times$ 1            & 10.7M & 19.5           & -              & 33.4           & -              \\
TRN~\cite{zhou2018temporal}  & 8     & 16G $\times$ N/A          & 18.3M & 34.4           & -              & 48.8           & -              \\
MFNet~\cite{lee2018motion}   & 10    & N/A $\times$ 10           & -     & 43.9           & 73.1           & -              & -              \\
CPNet~\cite{liu2019learning}   & 24    & N/A $\times$ 96           & -     & -              & -              & 57.7           & 84.0           \\
STM~\cite{jiang2019stm}       & 16    & 67G $\times$ 30  & 24.0M & 50.7           & 80.4           & 64.2           & 89.8 \\
TSM~\cite{lin2019tsm}       & 16+8  & 98G $\times$ 1   & 48.6M & 49.7           & 78.5           & 62.9           & 88.1           \\
TEA~\cite{li2020tea}     & 16       &  70G  $\times$ 1                        & 24.5M      & 51.9       &  80.3        & -          &  -              \\
TANet~\cite{liu2020tam} &16+8  & 99G $\times$ 1   & 25.6M & 50.6           & 79.3           & -           & -           \\
MSNet~\cite{kwon2020motionsqueeze}      & 16+8  & 101G $\times$ 10 & 49.2M & 55.1           & 84.0           & 67.1           & 91.0           \\
CIDC(R2D) ~\cite{li2020directional}  & 32    & 72G $\times$ 30   & 85M     & -           & -           & 40.2              & 68.6              \\
TEINet~\cite{liu2020teinet} &16+8  & 99G $\times$ 1   & 30.4M & 52.5           & -           & 65.5           & 89.8           \\
ACTION-Net~\cite{wang2021action} & 16  & 69.5G $\times$ 1            & 28.1M       & -        & - & 64.0   & 89.3 \\
TDN~\cite{wang2021tdn} & 16+8  & 198G $\times$ 1 & - & $56.8$           & 84.1           & $\mathbf{68.2}$           & 91.6           \\
\hline
TCM-R50          & 8      &     35G $\times$ 1                 &   24.5M    & 52.2      & 80.4        & 63.5           & 88.7                \\
TCM-R50          & 16     &     70G $\times$ 1          &   24.5M    & 53.1                & 81.2               &   65.1            & 89.6               \\
TCM-R50$_{En}$     & 16+8     &    105G $\times$ 1          &   49.0M    & 54.7                & 82.6                & 66.7       &  90.7              \\
TCM-R50$_{En}$     & 16+8     &    105G $\times$ 10          &   49.0M    & $\mathbf{57.2}$       & $\mathbf{85.2}$     &  $\mathbf{67.8}$    & $\mathbf{92.2}$ \\ \hline
\end{tabular}
\end{table*}

\begin{table*}
\caption{Performance comparison with the state-of-the-arts on the Kinetics-400 dataset. The symbol ``N/A” denotes the result that is not given.}
\label{tab:sotak}
\centering
\begin{tabular}{l|c|c|c|c|c}
\hline
Method                            & Pretrain   & Frame & FLOPs $\times$ Views   & \multicolumn{2}{c}{Kinetics-400}  \\
                                      &  &       &                            & Top-1(\%)          & Top-5(\%)           \\ \hline
ECO~\cite{zolfaghari2018eco}  & Scratch  & 92    & N/A $\times$ N/A             & 70.0         & 89.4                       \\
I3D~\cite{carreira2017quo}            &  Scratch      & 64       & 108G $\times$ N/A  & 72.1  & 90.3 \\
Two-Stream I3D~\cite{carreira2017quo} &  Scratch      & 64       & 216G $\times$ N/A  & 75.7  & 92.0 \\
R(2+1)D~\cite{tran2018closer}  & Sports-1M & 32     & 152G $\times$ 10    & 74.3           & 91.4          \\
ARTNet~\cite{wang2018appearance}  &  Scratch      & 16        & 23.5G $\times$ 250    & 69.2  & 88.3 \\
S3D-G~\cite{xie2018rethinking}  &   ImageNet       & 64       & 66.4G $\times$ N/A & 77.2  & 93.0 \\
Nonlocal-R50~\cite{NonLocal2018}       &   ImageNet       & 128       & 282G $\times$ 30  & 76.5  & 92.6 \\
Nonlocal-R101~\cite{NonLocal2018}       &   ImageNet       & 128       & 359G $\times$ 30       & 77.7  & 93.3 \\
ip-CSN~\cite{tran2019video}       &   ImageNet       & 32       & 83G $\times$ 30    & 76.7  & 92.3 \\
CIDC(I3D)~\cite{li2020directional}  & ImageNet & 32    & 92G $\times$ 30      & 74.5           & 91.3 \\
TPN~\cite{yang2020temporal}     &  Scratch      & 32        & N/A       & 78.9  & 93.9 \\
SmallBigNet~\cite{li2020smallbignet}  & Scratch       &32       &  418G $\times$ 12    & 77.4  & 93.3 \\
CorrNet~\cite{wang2020video}   &Scratch           & 32    & 224G $\times$ 30       & 79.2           & N/A           \\
SlowOnly~\cite{feichtenhofer2019slowfast}     &  Scratch     &8        & 41.9G $\times$ 30      & 74.8  & 91.6 \\
SlowFast~\cite{feichtenhofer2019slowfast}     &  Scratch      &8+32        & 106G $\times$ 30    & 77.9  & 93.2 \\
SlowFast~\cite{feichtenhofer2019slowfast}     &  Scratch      &16+64        & 234G $\times$ 30   & $\mathbf{79.8}$  & 93.9 \\
X3D~\cite{feichtenhofer2020x3d}       &  Scratch      &16        & 48.4G $\times$ 30      & 79.1  & 93.9 \\
\hline
TSN~\cite{wang2016temporal}  & ImageNet & 25     & 3.2G $\times$ 10       & 72.5           & 90.5          \\

TSM~\cite{lin2019tsm}      & ImageNet & 16  & 65G $\times$ 30    & 74.7           & 91.4          \\
STM~\cite{jiang2019stm}      & ImageNet & 16    & 67G $\times$ 30  & 73.7           & 91.6    \\
TEA~\cite{li2020tea}    & ImageNet & 16       &  70G  $\times$ 30    & 76.1       &  92.5      \\
MSNet~\cite{kwon2020motionsqueeze}     & ImageNet & 16  & 67G $\times$ 30  & 76.4           & N/A \\
CIDC(R2D)~\cite{li2020directional}  &ImageNet & 32    & 72G $\times$ 30   & 72.2           & 90.1 \\
TEINet~\cite{liu2020teinet} & ImageNet &16  & 66G $\times$ 30    & 76.2           & 92.5           \\
TANet-152~\cite{liu2020tam} & ImageNet &16  & 242G $\times$ 12   & 79.3           & 94.1           \\
TDN~\cite{wang2021tdn}    & ImageNet  & 16+8  & 198G $\times$ 30  & 79.4           & $\mathbf{94.4}$  \\
\hline
TCM-R50        & ImageNet & 8      &     35G $\times$ 30           & 76.1     & 92.3       \\
TCM-R50        & ImageNet & 16     &     70G $\times$ 30            & 77.4     & 93.1               \\
TCM-R50$_{En}$   & ImageNet & 16+8     &    105G $\times$ 30      & $\mathbf{78.5}$       & $\mathbf{93.8}$     \\
\hline
\end{tabular}
\end{table*}

\subsection{Comparison with state-of-the-arts}
To evaluate the temporal modeling ability and the entire capacity of our method, we compare our TCM-derived models with the state-of-the-arts extensively on both the temporal dominated datasets Something-Something V1 \& V2 and the scene dominated datasets Kinetics-400, UCF-101 and HMDB-51. The results are reported in Table~\ref{tab:sota}, Table~\ref{tab:sotak} and Table~\ref{tab:ucf-hmdb}.

Table \ref{tab:sota} shows the performance of 23 recent action recognition methods on the Something-Something V1 \& V2 datasets. There are three parts in this table: 3D CNN methods~\cite{zolfaghari2018eco,wang2018videos,xie2018rethinking,martinez2019action} (in the upper part), 2D CNN methods~\cite{wang2016temporal,zhou2018temporal,lee2018motion,liu2019learning,jiang2019stm,li2020tea,kwon2020motionsqueeze} (in the middle part), and the proposed TCM based methods (in the bottom part). Without any bells and whistles, the Top-1 accuracy of our ``TSM+TCM" method \textit{i.e.} ``TCM-R50" on the Something-Something V1 dataset reaches to 52.2\%, which surpasses its 2D CNN based counterparts STM~\cite{jiang2019stm}, TEA~\cite{li2020tea}, and TANet~\cite{liu2020tam} that need double input (16 input frames) for at least 0.3\%. Importantly, with 16 input frames, the accuracy of our ``TSM+TCM" method on the Something-Something V1 \& V2 datasets is further improved, where its Top-1 accuracy is higher than STM~\cite{jiang2019stm}, TEA~\cite{li2020tea}, and TANet~\cite{liu2020tam} for at least 1.2\% on Something-Something V1 and 0.9\% on Something-Something V2. In addition, compared to the 3D CNN based methods, \textit{e.g.} 3D DenseNet121~\cite{zhou2020spatiotemporal}, the Top-1 accuracy is improved by 2.9\% (53.1\% vs. 50.2\%) and 2.2\% (65.1\% vs. 62.9\%) separately on the Something-Something V1 dataset and the Something-Something V2 dataset. Compared to NL-I3D~\cite{wang2018videos}, the performance is improved by 7.8\% with using less input (8 vs. 32) and less computation (35GFlops vs. 168GFlops $\times$ 2).

Following~\cite{lin2019tsm,kwon2020motionsqueeze}, we ensemble our 8-frame and 16-frame models by averaging their prediction scores.
Our 10-clip model obtains remarkable results on the Something-Something V1 dataset. Specifically, as shown in the last row of Table \ref{tab:sota}, its Top-1 and Top-5 accuracy outperforms the state-of-the-art method TDN~\cite{wang2021tdn} by 0.4\% (57.2\% vs. 56.8\%) and 1.1\% (85.2\% vs. 84.1\%), respectively. Furthermore, on the Something-Something V2 dataset, in contrast to TDN~\cite{wang2021tdn}, our 10-clip model also has a better performance in the Top-5 accuracy (92.2\% vs. 91.6\%) and its Top-1 accuracy is just a little bit lower (67.8\% vs. 68.2\%).

Table \ref{tab:sotak} shows the comparison with the state-of-the-art approaches on the scene dominated dataset Kinetics-400. It can be clearly seen that our TCM has an outstanding performance. Firstly, our 8-frame TCM-R50 surpasses the 64-frame I3D method~\cite{carreira2017quo} (Top-1 accuracy: 76.1\% vs. 72.1\%), and it achieves a competitive accuracy to the 128-frame Nonlocal-R50 approach~\cite{NonLocal2018} (Top-1 accuracy: 76.1\% vs. 76.5\%) while its GFlops is 8$\times$ less. Moreover, our 8-frame TCM-R50 model performs even better than the 8-frame SlowOnly method~\cite{feichtenhofer2019slowfast} (Top-1 accuracy: 76.1\% vs. 74.8\%) with 1.2$\times$ less GFlops. All these results demonstrate that, our TCM network, is more accurate and efficient than the non-local network to model the temporal relationships for video classification. Secondly, our 16-frame TCM-R50 outperforms most of its 16-frame counterparts, \textit{i.e.} STM~\cite{jiang2019stm} (Top-1 accuracy: 77.4\% vs. 73.7\%), TEA~\cite{li2020tea} (Top-1 accuracy: 77.4\% vs. 76.1\%), MSNet~\cite{kwon2020motionsqueeze} (Top-1 accuracy:77.4\% vs. 76.4\%), and TEINet~\cite{liu2020teinet} (Top-1 accuracy: 77.4\% vs. 76.2\%). With 5.1$\times$ less GFlops than the 128-frame Nonlocal-R101 method~\cite{NonLocal2018}, our model obtains a comparable accuracy (Top1 acc: 77.4\% vs. 77.7\%). Thirdly, we perform the score fusion over 8-frame TCM-R50 and 16-frame TCM-R50, which mimics the two-steam fusion with two temporal rates. At testing, we use 10 clips and 3 crops per clip. Our TCM achieves a higher accuracy than the ``8+32-frame" SlowFast model~\cite{feichtenhofer2019slowfast}, with using less input frames and a little bit less GFlops (105 vs. 106). Table \ref{tab:sotak} reveals that, spatio-temporal learning of our TCM is more effective than temporal shift of TSM.

\setlength{\tabcolsep}{4.5pt}
\begin{table}
\caption{Comparison with the state-of-the-arts on the UCF-101 and HMDB-51 datasets. The symbol ``-” denotes the result that is not given.}
\label{tab:ucf-hmdb}
\centering
\begin{tabular}{l|c|c|c|c}
\hline
Method & Pretrain & Backbone & UCF-101 & HMDB-51 \\
\hline TSN~\cite{wang2016temporal} & ImageNet & Inception V2 & $86.4 \%$ & $53.7 \%$ \\
P3D~\cite{qiu2017learning} & ImageNet & ResNet50 & $88.6 \%$ & $-$ \\
C3D~\cite{tran2015learning} & Sports-1M & ResNet18 & $85.8 \%$ & $54.9 \%$ \\
I3D~\cite{carreira2017quo} & Kinetics & Inception V2 & $95.6 \%$ & $74.8 \%$ \\
ARTNet~\cite{wang2018appearance} & Kinetics & ResNet18 & $94.3 \%$ & $70.9 \%$ \\
S3D~\cite{xie2018rethinking} & Kinetics & Inception V2 & $96.8 \%$ & $75.9 \%$ \\
R(2+1)D~\cite{tran2018closer} & Kinetics & ResNet34 & $96.8 \%$ & $74.5 \%$ \\
TSM~\cite{lin2019tsm} & Kinetics & ResNet50 & $96.0 \%$ & $73.2 \%$ \\
STM~\cite{jiang2019stm} & Kinetics & ResNet50 & $96.2 \%$ & $72.2 \%$ \\
TEA~\cite{li2020tea} & Kinetics & ResNet50 & $96.9 \%$ & $73.3 \%$ \\
TDN~\cite{wang2021tdn} & Kinetics & ResNet50 & $\mathbf{97.4\%}$ & $76.3\%$ \\ \hline
TCM-R50 (ours) & Kinetics & ResNet50 & $\mathbf{97.1\%}$ & $\mathbf{77.5\%}$ \\
\hline
\end{tabular}
\end{table}

To further verify the generalization ability of the explored TCM, we transfer the trained 16-frame TCM-R50 model from the Kinetics-400 dataset to the UCF-101 and HMDB-51 datasets same as the previous works~\cite{lin2019tsm, li2020tea, wang2021tdn}. We follow the standard evaluation metric on the two datasets and report the average Top-1 accuracy over the three splits, where the results are summarized in Table~\ref{tab:ucf-hmdb}. We compare our TCM with the advanced methods such as the 2D baseline TSM~\cite{lin2019tsm}, the 3D CNN based methods I3D~\cite{carreira2017quo}, C3D~\cite{tran2015learning}, and R(2+1)D~\cite{tran2018closer}, and the other temporal modeling methods~\cite{jiang2019stm,li2020tea,wang2021tdn}. From the results, we can see that our TCM is superior to these methods, and the performance improvement is more obvious on the HMDB51 dataset which is boosted by at leat 1.2\% (Top-1 accuracy: 77.5\% vs. 76.3\%). The human actions in HMDB51 are more relevant with motion information, therefore temporal modeling is more important on this dataset. On the other side, on the UCF-101 dataset, our TCM-R50 also achieves competitive result to the first place (Top-1 accuracy: 97.1\% vs. 97.4\%).

\subsection{Comparison with TPN}
To make a fair comparison, we compare the performance of our TCM and TPN~\cite{yang2020temporal} on the Something-Something V1 \& V2 datasets, using the same backbone (TSM-R50~\cite{lin2019tsm}) and same experiment settings. Results in Table~\ref{tab:tpn_tcm} demonstrate that TCM is superior to TPN: 1) Higher accuracy, where it outperforms TPN on the Something-Something V1 dataset by 3.2\% and the Something-Something V2 dataset by 1.5\%; 2) Fewer parameters, where its parameters is less than 30\% of TPN (TCM vs. TPN: 24.5M vs. 82.5M). 3) Less computation, where it takes only about 85\% Flops of TPN ( TCM vs. TPN: 35.3G vs. 41.5G). The efficiency of TPN is not satisfactory may due to the use of 3D convolution. TCM is more efficient since it is based on 2D convolution.

\begin{table}
\caption{Performance comparison with TPN on the Something-Something V1 \& V2 datasets. Specifically, 8 frames are input for training. Single-clip and
center-crop testing scheme is used here.}
\label{tab:tpn_tcm}
\centering
\begin{tabular}{|l|c|c|c|c|c}
\hline
Method                         & FLOPs      & Params & Top1@V1 &  Top1@V2 \\ \hline
TSM-R50~\cite{lin2019tsm}        & 33.4G   & 24.3M   & 45.6  & 59.1            \\ \hline
TSM+TPN~\cite{yang2020temporal}    & 41.5G   & 82.5M     & 49.0      & 62.0    \\ \hline
TSM+TCM(ours)    &     35.3G     &   24.5M    & \textbf{52.2}       & \textbf{63.5}      \\ \hline
\end{tabular}
\end{table}

\subsection{Ablation Study}
Ablation studies about the components of our TCM are conducted on the Something-Something V1 dataset. Particularly, the ResNet-18 with the temporal shift module~\cite{lin2019tsm} is served as the backbone here. Following the setting of~\cite{lin2019tsm}, 8 input frames, which are sampled from the video via the segment-based sampling method~\cite{wang2016temporal}, are utilized for training and inference. The training parameters are: the training epochs are 40, the batch size is 64, the initial learning rate is 0.02 (decays by 0.1 at epoch $20~\&~30$), the weight decay is 5e-4, and the dropout is 0.5.

\subsubsection{Which feature source is most suitable for building a multi-scale temporal motion pyramid?}
Extensive experiments on feature sources are tested, results shown in Table \ref{tab:featuresource} verify that the proposed TCM overcomes the previous approach's inability to explore benefits from relatively shallow sources, e.g. \textit{res2} or \textit{res3}. Even for high-level feature sources, significant improvement is gained. Enjoying the flexibility to plug-and-play in a single-layer, the multi-layer style can also be performed by using multiple TCM modules simultaneously. Its performance is slightly improved compared to the baseline but is degraded in contrast to the usage of a single TCM. This is because stacking multiple TCM modules damages the brightness consistency of the prior TCM layer. Since \textit{res2} is too shallow to extract enough spatial features, the accuracy increases the most when TCM is placed right after the \textit{res3} layer. As a result, we select to use a single TCM right behind layer \textit{res3} finally.
\begin{table}[!htbp]
\centering
\caption{Performance comparison with different feature sources. TCM is placed right behind the specified layer.}
\label{tab:featuresource}
\begin{tabular}{l|c|c|c}
\hline Layer & GFLOPs & Top-1(\%) & Top-5(\%) \\
\hline baseline & $14.5$ & 42.8 & 72.3 \\ \hline
res2 & $17.5$ & 42.9 & 72.6 \\
res3 & $16.4$ & $\mathbf{45.9}$ & $\mathbf{75.7}$ \\
res4 & $15.1$ & 43.8 & 73.1 \\
res5 & $14.7$ & 43.6 & 72.8 \\
res\{2,3\} & $19.4$ & 45.8 & 75.6 \\
res\{2,3,4\} & $20.0$ & 45.8 & 75.5 \\
res\{2,3,4,5\} & $20.2$ & 45.6 & 75.4 \\
\hline
\end{tabular}
\end{table}

\subsubsection{How important of MTDM and TAM?}
MTDM is used for multi-scale temporal dynamics extraction and TAM is used for temporal scales aggregation. The effect of these modules is studied in Table \ref{tab:mtdm-tam}. From which we can observe: both MTDM and TAM can enhance the accuracy of action recognition largely, and the action recognition performance is further boosted when combining MTDM with TAM.
\begin{table}[!ht]
\centering
\caption{Performance comparison with different TCM components.}
\label{tab:mtdm-tam}
\begin{tabular}{ccc|c}
\hline MTDM & TAM &  Top-1(\%) & $\Delta$ Top-1(\%) \\ \hline
 & & 42.6 & baseline \\ \hline
 $\checkmark$ & & 43.8 & +1.2 \\
  & $\checkmark$ & 43.2 & +0.5 \\
 $\checkmark$ & $\checkmark$ & $\mathbf{45.9}$ & +3.3 \\ \hline
\end{tabular}
\end{table}

\subsubsection{What is the difference with the existing motion cue learning method?}
As far as we known, the existing motion cue learning methods~\cite{fan2018end,kwon2020motionsqueeze,piergiovanni2019representation,jiang2019stm,wang2020video} mainly aimed at extracting the feature-level motion patterns between adjacent frames. TVNet~\cite{fan2018end} and Rep-Flow~\cite{fan2018end,piergiovanni2019representation} internalize the TV-L1 optical flow in their networks, which enabling to capture the motion information and appearance information in an end-to-end way. STM~\cite{jiang2019stm}, TEA~\cite{li2020tea}, and CorrNet~\cite{wang2020video} propose approaches to establish frame-to-frame matches over convolutional feature maps, and well-designed blocks for learning better temporal information are applied to replace the original residual blocks in the ResNet architecture to construct a simple yet effective network. MSNet~\cite{kwon2020motionsqueeze} presents a trainable module named MotionSqueeze to substitute the external and heavy computational optical flow with the internal and lightweight learned motion features. These works have promoted video understanding, but the extraction of action visual tempo from video sequence features has been rarely accounted. Our work aims to fill this gap, where the video sequence features are exploited to capture the movement patterns on different temporal scales, and the useful visual tempo information is enhanced adaptively. We compared our method with these motion cue learning methods on the Kinetics-400 dataset in Table \ref{tab:motion-k400}, it reveals that our method is superior to these methods due to it concerns multi-scale fine-grained temporal dynamics of both the fast-tempo and the slow-tempo.
\begin{table}[!ht]
\centering
\caption{Performance comparison with the existing motion cue learning methods on the Kinetics-400 dataset. The symbol ``-” denotes the result that is not given.}
\label{tab:motion-k400}
\begin{threeparttable}
\begin{tabular}{lcccc}
\hline
Method                   & Frame & GFLOPs $\times$ Clips & FPS & Top-1(\%) \\ \hline
TVNet~\cite{fan2018end} & 18    & N/A $\times$ 250     & -    & 68.5$^1$ \\
Rep-flow~\cite{piergiovanni2019representation} & 32    & 152$^{2} \times$ 25     & 2.0    & 75.5 \\
STM~\cite{jiang2019stm} & 16    & 67$\times$ 30      & -    & 73.7 \\
TEA~\cite{li2020tea}    & 16    & 70$\times$ 1       & -    & 74.0 \\
CorrNet-50~\cite{wang2020video} & 32  & 115 $\times$ 10      & -    & 77.2 \\
MSNet~\cite{kwon2020motionsqueeze}  & 16    & 67 $\times$ 10     & 31.2  & $76.4$ \\
TCM(ours)                     & 16    & 70$\times$ 10     & 28.1  & $\mathbf{77.2}$ \\
\hline
\end{tabular}
\begin{tablenotes}
  \footnotesize
  \item[1] represents that the results of the method are reproduced by us according to the source code they supplied.
\end{tablenotes}
\end{threeparttable}
\end{table}

\begin{table}[!htbp]
\caption{Performance comparison with the plug-in-and-play modules on the Something-Something V2 dataset (TSM-R50 is used as a baseline). All methods use ResNet50 pre-trained on ImageNet as the backbone and 8-frame input for fair comparison. The least Param. and the best result are highlighted as bold. The symbol ``-” denotes the result that is not given.}
\label{tab:plugin}
\centering
\begin{tabular}{lcccccccc}
\hline
Method                           & GFLOPs & Param.           & \multicolumn{2}{c}{Sth-Sth V2}  \\
                                 &       &         & Top-1(\%)          & Top-5(\%)    \\ \hline
baseline(TSM-R50~\cite{lin2019tsm})            & 32.8    & 24.3M   & 58.8           & 85.4   \\ \hline
Nonlocal TSM-R50~\cite{lin2019tsm}           & 49.3    & 31.2M   & 60.1           & 86.7   \\
TEINet~\cite{liu2020teinet}      & 33  & 30.4M   & 61.3           & -                \\
MSNet~\cite{kwon2020motionsqueeze} & 34.3  & 24.6M   & 63.0   & 88.4 \\
ACTION-Net~\cite{wang2021action} & 34.7  & 28.1M   & 62.5   & 87.3 \\
TCM-R50(ours)  &     35.3  &   $\mathbf{24.5M}$        & $\mathbf{63.5}$           & $\mathbf{88.7}$                \\ \hline
\end{tabular}
\end{table}

\subsubsection{Compared with plug-in-and-play modules.}
We make a comprehensive comparison with methods~\cite{lin2019tsm,liu2020teinet,kwon2020motionsqueeze,wang2021action}, which enjoy a plug-and-play manner likes our TCM, on the Something-Something V2 dataset, the results are shown in Table \ref{tab:plugin}. The TSM-R50~\cite{lin2019tsm} method is served as the baseline here for performance and efficiency comparison. For the Nonlocal TSM-R50 method~\cite{lin2019tsm}, we retrained the model on the Something-Something V2 dataset via the official PyTorch code base~\cite{lin2019tsm}. Compared to Nonlocal TSM-R50, our TCM-R50 uses only ~0.7$\times$ GFLops and ~0.8$\times$vParam., and achieves much better performance (Top1 accuracy: 63.5\% vs. 60.1\%, Top5 accuracy: 88.7\% vs. 85.4\%). Probably because the non-local module focuses on capturing global dependencies, in contrast, our approach pays close attention to extract the most appropriate dependencies. In the spatial dimension, pixel-wise optical-flow-like motion information is obtained by an optimized MotionSqueeze~\cite{kwon2020motionsqueeze} algorithm. In the temporal dimension, for the first time, we sample the learned features with different ratios, and carry out an efficient temporal attention consideration that involves cross-temporal interaction. Compared to other well-designed approaches, our TCM-R50 utilizes the fewest parameters and gets the highest performance improvement.

\subsection{Visualization}
\begin{figure*}[!tbp]
  \includegraphics[width=1.02\textwidth]{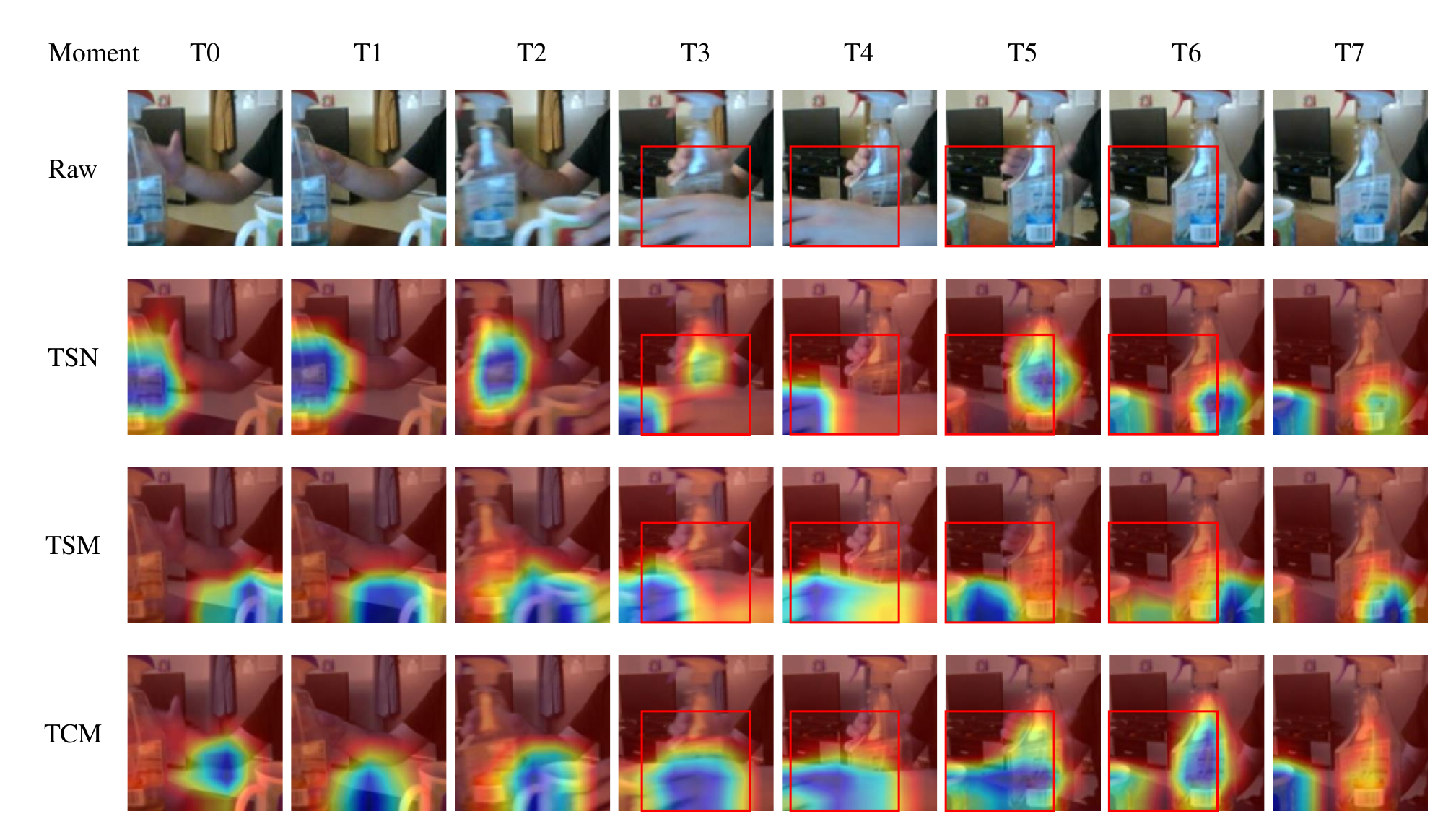}
  \caption{Visualization of activation maps with Grad-CAM~\cite{jacobgilpytorchcam} for the deep features of the action ``Moving something and something so they pass each other", where the deep features are extracted by TSN, TSM and our TCM methods, respectively. All methods use 8-frame input to visualize on the Something-Something V1 dataset. In the first row, we plot the 8 RGB raw frames, then we plot the activation maps of TSN, TSM and our TCM. We use red bounding box to highlight the regions that need to be focused on in the activation maps. Compared to TSN and TSM, it can be noticed that TCM is able to learn the deep features related to human interaction with objects \textit{i.e. the red bounding boxes at moment T3-T6 frames.} Particularly, in contrast to TSM, we can find that the exploited TCM has the following advantages: 1) locating the area where the cup and bottle pass through each other more accurate (\textit{i.e.} frames at moment T3), 2) depicting the interaction of the hand and cup more precise (\textit{i.e.} frames at moment T4), 3) considering the connection between the cup and the bottle (\textit{i.e.} frames at moment T5 and T6).}
  \label{fig:gradcam}
\end{figure*}

To further explore the working mechanism of our TCM, we visualize the class activation maps with Grad-CAM~\cite{zhou2016learning,selvaraju2017grad,jacobgilpytorchcam}. Fig. \ref{fig:gradcam} shows the feature visualizations that are characterized by TSN~\cite{wang2016temporal}, TSM~\cite{lin2019tsm} and our TCM for the action ``Moving something and something so they pass each other" on the Something-Something V1 dataset. In our visualization, we take 8 frames from the moment T0 to the moment T7 as input, and plot the activation maps for each frame. From the results, it can be found that 1) TSN only pays attention to objects and fails to capture the movement of objects and human hands; 2) TSM can capture the coarse motion information, but it is unable to accurately locate the action-relevant regions; 3) Our TCM is superior to the TSM baseline in focusing on the action-relevant regions, due to the long-term and short-term temporal modeling capacity of the proposed TCM.

\begin{figure}[!htbp]
  \includegraphics[width=0.5\textwidth]{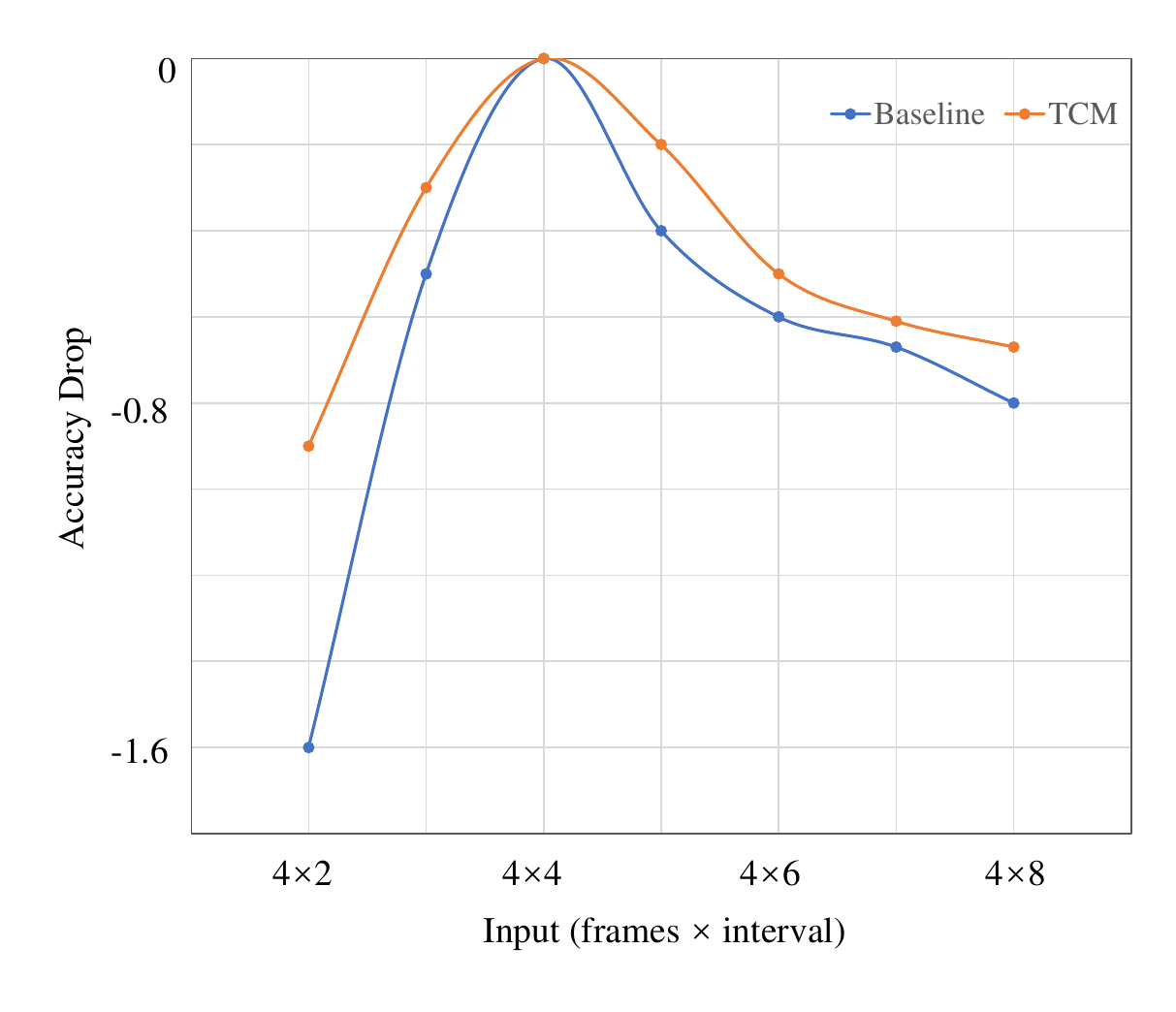}
  \caption{Robustness to the variance of action visual tempo. The orange line depicts the accuracy change of the baseline network with incorporating our TCM, while the blue line denotes the accuracy change of the baseline without integrating our TCM.}
  \label{fig:drop}
\end{figure}

\subsection{Empirical Analysis}
To study the robustness of TCM to action visual tempo variation, we follow~\cite{yang2020temporal} to evaluate the accuracy drop by re-sampling the input frames with different temporal intervals. We first train TSM-ResNet18 and TCM-ResNet18 on the Something-Something V1 dataset with $4 \times 4$ (frames $\times$ interval) inputs, then we re-scale the original $4\times4$ input by re-sampling the frames with the stride $\tau$ equals to \{2,3,4,5,6,7,8\} respectively, accordingly the temporal scales of a given action instance are adjusted. For some videos with insufficient number of frames, we copy the last frame until the number of the input frames is reached. Fig. \ref{fig:drop} shows the accuracy curves of varying the action temporal scales for TSM-ResNet18 and our TCM-ResNet18. Clearly, our TCM improves the robustness of the baseline, and gets a smoother curve (see the orange curve on the top), which strongly supports the conclusion that our TCM can effectively extract and fuse the action visual tempo features.

\section{Conclusion}
We propose a novel Temporal Correlation Module (TCM) to deal with the variation of action visual tempo in videos, which includes a Multi-scale Temporal Dynamics Module (MTDM) and a Temporal Attention Module (TAM). MTDM extracts pixel-wise fine-grained temporal dynamics for both the fast-tempo and the slow-tempo by utilizing a correlation operation. TAM adaptively selects and enhances the most effective action visual tempo information by taking across-temporal dynamics interaction into account. The explored TCM can be seamless integrated into the current action recognition backbones and optimized in an end-to-end way to capture the action visual tempo commendably. It is especially effective when incorporating it to the low-level layer of the backbone. Extensive experiments on 5 representative datasets have demonstrated the effectiveness of TCM in both accuracy and efficiency.

\section*{Acknowledgment}
This work was supported by the National Natural Science Foundation of China under Grant 62106177. It was also supported by the Central University Basic Research Fund of China (No.2042020KF0016). The numerical calculation was supported by the supercomputing system in the Super-computing Center of Wuhan University.

\ifCLASSOPTIONcaptionsoff
  \newpage
\fi

\bibliographystyle{IEEEtran}
\bibliography{main.bib}

\begin{IEEEbiography}[{\includegraphics[width=1in,height=1.25in,clip,keepaspectratio]{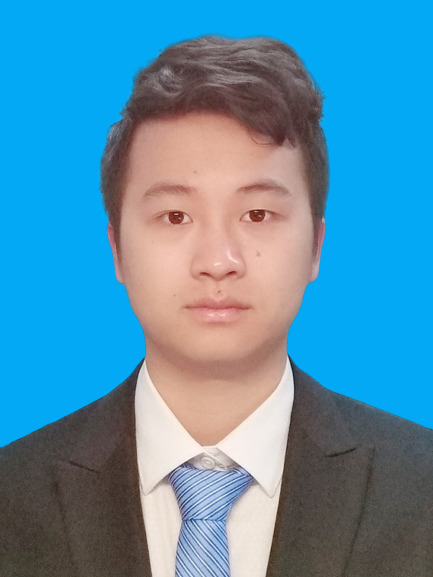}}]{Yuanzhong Liu}
Yuanzhong Liu is currently a postgraduate student at Wuhan University. He received the B. Eng. degree from the school of resources and environment from University of Electronic Science and Technology of China in 2020. His research interests mainly include computer vision and machine learning.
\end{IEEEbiography}

\vspace{-1.0cm}

\begin{IEEEbiography}[{\includegraphics[width=0.95\textwidth]{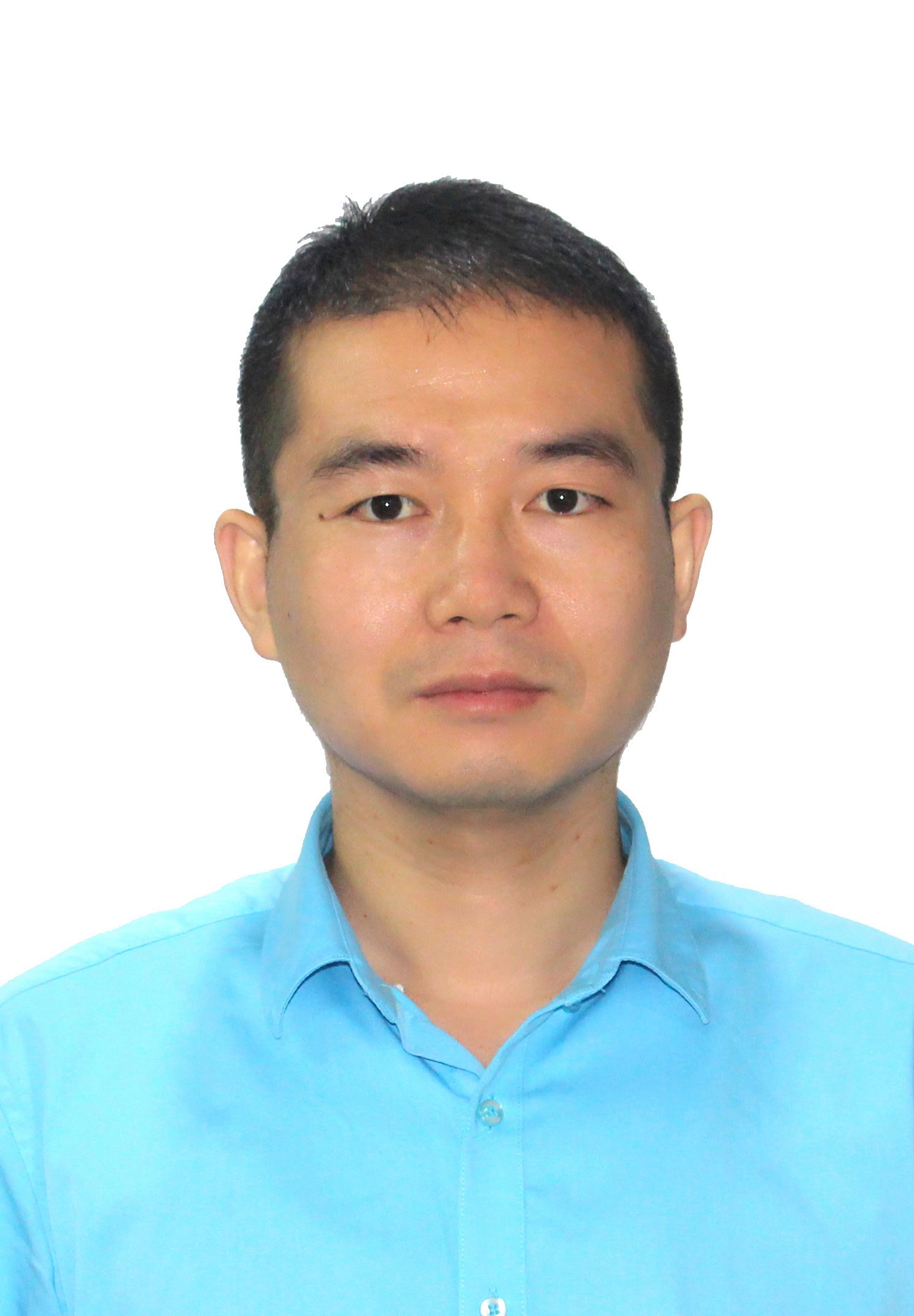}}]
{\textbf{Zhigang Tu}} started his Master Degree in image processing at Wuhan University, China, 2008. In 2015, he received the Ph.D. degree in Computer Science from Utrecht University, Netherlands. From 2015 to 2016, he was a postdoctoral researcher at Arizona State University, US. Then from 2016 to 2018, he was a research fellow at Nanyang Technological University, Singapore. He is currently a professor at the State Key Laboratory of Information Engineering in Surveying, Mapping and Remote sensing, Wuhan University. His research interests include computer vision, image processing, video analytics, and machine learning. Special for motion estimation, video super-resolution, object segmentation, action recognition and localization, and anomaly event detection. 

He has co-/authored more than 50 articles on international SCI-indexed journals and conferences. He is an Associate Editor of the SCI-indexed journal \textit{The Visual Computer} (IF=2.601), a Guest Editor of \textit{Journal of Visual Communications and Image Representation} (IF=2.678) and \textit{CC\&HTS} (IF=1.339). He is the first organizer of the ACCV2020 Workshop on MMHAU (Japan). He received the ``Best Student Paper" Award in the $4^{th}$ Asian Conference on Artificial Intelligence Technology.
\end{IEEEbiography}

\vspace{-1.0cm}

\begin{IEEEbiography}[{\includegraphics[width=0.95\textwidth]{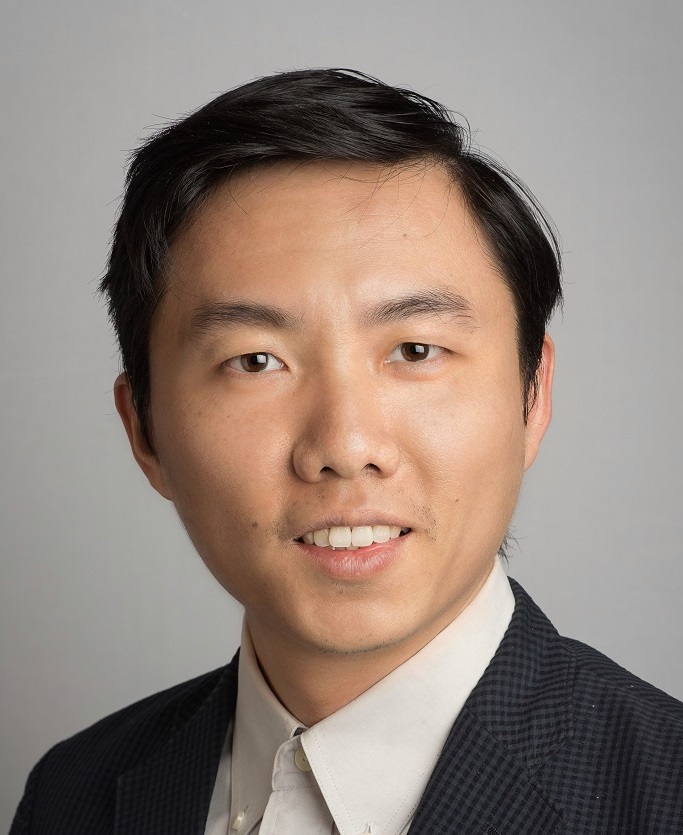}}]
{\textbf{Junsong Yuan}} is Professor and Director of Visual Computing Lab at Department of Computer Science and Engineering, State University of New York at Buffalo (UB), USA. Before that he was Associate Professor (2015-2018) and Nanyang Assistant Professor (2009-2015) at Nanyang Technological University (NTU), Singapore. He obtained his Ph.D. from Northwestern University in 2009, M. Eng. from National University of Singapore in 2005, and B. Eng. from Huazhong University of Science Technology in 2002. His research interests include computer vision, pattern recognition, video analytics, human action and gesture analysis, large-scale visual search and mining. He received Best Paper Award from IEEE Trans. on Multimedia, Nanyang Assistant Professorship from NTU, and Outstanding EECS Ph.D. Thesis award from Northwestern University. 

He served as Associate Editor of IEEE Trans. on Image Processing (T-IP), IEEE Trans. on Circuits and Systems for Video Technology (T-CSVT), Machine Vision and Applications, and Senior Area Editor of Journal of Visual Communications and Image Representation. He was Program Co-Chair of IEEE Conf. on Multimedia Expo (ICME'18), and Area Chair for CVPR, ICCV, ECCV, and ACM MM. He was elected senator at both NTU and UB. He is a Fellow of IEEE and IAPR.
\end{IEEEbiography}

\end{document}